# Update rules for parameter estimation in Bayesian networks

Eric Bauer
Stanford University
ebauer@cs.stanford.edu

Daphne Koller
Stanford University
koller@cs.stanford.edu

Yoram Singer
AT&T Labs
singer@research.att.com

## Abstract

This paper re-examines the problem of parameter estimation in Bayesian networks with missing values and hidden variables from the perspective of recent work in on-line learning [13]. We provide a unified framework for parameter estimation that encompasses both on-line learning, where the model is continuously adapted to new data cases as they arrive, and the more traditional batch learning, where a pre-accumulated set of samples is used in a one-time model selection process. In the batch case, our framework encompasses both the gradient projection algorithm [2, 3] and the EM algorithm [15] for Bayesian networks. The framework also leads to new on-line and batch parameter update schemes, including a parameterized version of EM. We provide both empirical and theoretical results indicating that parameterized EM allows faster convergence to the maximum likelihood parameters than does standard EM.

## 1 Introduction

Over the past few years, there has been a growing interest in the problem of learning Bayesian networks from data. The reasons behind this trend are clear. Manually engineering a large Bayesian network is a difficult and time-consuming process. Furthermore, it is never clear whether the network designed by the expert is really the most appropriate model for the domain. Finally, the world is not always static; we want our model to adapt itself automatically to changing conditions.

So far, most of the work on learning Bayesian networks has been devoted to the *batch learning* task: we are given a training set consisting of some number of data cases, and our task is to construct a Bayesian network which best models the data. We can also consider the problem from the somewhat different perspective of *on-line learning*. In this task, we assume that we have an existing model which we must adapt to some data $D$, thus forcing us to balance the desire to improve our model's fit to $D$ with the desire to maintain information already included in the model. On-line learning is designed to deal with situations where we want to fine-tune an existing model, either because the model was initially inaccurate or because the environment has changed.

We derive a precise formulation for these tasks using the new framework introduced by Kivinen and Warmuth [13]. Our analysis leads to the definition of a family of new algorithms for the traditional task of batch parameter estimation for Bayesian networks. We show, both theoretically and empirically, that the convergence properties of these new algorithms can be significantly better than those of the current state-of-the-art algorithms for this problem such as EM [15, 9]. We then use the same framework to provide an initial foundation for on-line parameter estimation for Bayesian networks.

In many real-world domains, the data available for learning is incomplete: some of the variables may be difficult or even impossible to observe. It is therefore important that our learning algorithm be able to make effective use of partially specified data cases. This need is particularly crucial in the context of on-line learning, where the primary source of data cases lies within the queries presented to the network for inference. Clearly, there is little point in presenting to the network queries about data cases where all variables have already been observed. Thus, we focus on the problem of learning from a data set consisting of data cases where some of the variables may be permanently or occasionally unobserved.

In the presence of missing data, the problem of Bayesian network learning becomes much more difficult [9]. In fact, even the relatively simple task of adapting the numerical parameters (conditional probability table entries) for a given network structure becomes nontrivial. Parameter estimation is an important task both because of the difficulty of eliciting accurate numerical estimates from people, and because parameter learning is part of the inner loop of more general algorithms that also learn the structure (see [9] for a survey). In this paper, we choose to focus on this aspect of the learning task, deferring treatment of structure learning to future work. (See [8] for some preliminary work on this problem.)

Formally, we are given some Bayesian network $B$, described by a vector $\bar{\theta}$ of numerical parameters, and a set $D$ of one or more data cases. We want to use $D$ to construct a new network $\tilde{B}$, described by a new parameter vector $\tilde{\theta}$. As we will see, this formalization applies equally well to the batch learning task and the on-line learning task.

Two factors guide the choice of $\tilde{\theta}$: the extent to which it fits $D$, and the fact that we don't want to move too far away

4    Bauer, Koller, and Singer

from our existing model $\bar{\theta}$. In Section 2, we formalize this intuition by optimizing a function $F$ which incorporates both the log-likelihood of $D$ and the distance between $\bar{\theta}$ and $\tilde{\theta}$. An analysis of the optimal value for this function results in an *update rule*, which tells us how $\tilde{\theta}$ can be constructed from $\bar{\theta}$.

The exact form of the function $F$ depends on our choice of the distance function used to compare $\bar{\theta}$ and $\tilde{\theta}$, and on the relative weight we give the distance and the log-likelihood. Each choice leads to a different update rule. We perform the analysis for various choices of distance measure, including $\mathcal{L}_2$-norm, relative entropy (KL-divergence), and $\chi^2$-distance (a linear approximation to relative entropy).

In Section 3, we instantiate our update rules from Section 2 to the batch learning context. In this case, we typically construct a model by iterating through the same data set $D$ multiple times. Then, $\bar{\theta}$ is the model constructed in the previous iteration, and $\tilde{\theta}$ is the result of another pass of adapting it to the data. Surprisingly, it turns out that well-known parameter estimation algorithms are special cases of our framework. For example, our update rule for $\mathcal{L}_2$-norm results in the gradient-projection scheme of [3, 22, 2]. Our update rule for the $\chi^2$ distance results in a family of update rules with varying learning rates $\eta$. This family, which we denote EM($\eta$), includes the standard EM algorithm [6, 9] as the special case EM(1). From the relative entropy distance, we derive an analogous family of multiplicative update rules which, following [13, 11], we call EG($\eta$).

We provide both theoretical and empirical evidence showing that EM($\eta$) can lead to much faster convergence than standard EM while still using a very simple update rule. (In contrast to more complex and expensive second order methods such as those of [23]). In particular, we show in Section 4 that, while 1 is the largest value of $\eta$ for which convergence to a local maximum is guaranteed, some value $\eta^*$ which is bigger than 1 provides the optimal convergence rate. More precisely, for any (local or global) maximum of the likelihood function, there is a value $\eta^* > 1$ and a neighborhood around the local maximum, such that EM($\eta^*$) provides the fastest convergence (of any EM($\eta$) algorithm) to the maximum in that neighborhood.

While the optimal value $\eta^*$ cannot be computed, evidence acquired when applying EM($\eta$) algorithms to other learning tasks shows that certain values of $\eta$ seem to work well for most problems. In Section 5, we provide experiments with EM($\eta$) for various values of $\eta$. These experiments indicate that a value of $\eta = 1.8$ appears to work well, and certainly much better than $\eta = 1$. In particular, we show that EM(1.8) often requires approximately half the number of iterations to converge. Since each iteration of a parameter estimation algorithm (whether EM or EM($\eta$)) involves a Bayesian network computation for each instance in our data set, the computational savings resulting from a large reduction in the number of iterations can be very significant.

Finally, we return in Section 6 to the problem of on-line learning. In this case, our basic framework of Section 2 should be interpreted as applying to a $D$ consisting of a single new sample, and the current model $\bar{\theta}$ is simply the one constructed based on the samples seen so far. We examine the resulting update rules, and discuss their applicability to the problem of gradually adapting network parameters over the entire lifetime of a Bayesian network. We conclude in Section 7 with some discussion and open questions.

## 2    The framework

In this section, we present a basic framework which can be used to interpret both the on-line learning and the more standard batch learning tasks. Our presentation and notation follow that of [9].

Recall that our task is to learn the numerical parameters for a discrete valued Bayesian network of a given structure $S$. Most simply, a Bayesian network is parameterized using a vector of *conditional probability table (CPT) entries*, one entry for each value of each node and each instantiation of the nodes parents. More precisely, let $X_i$ be a node in the network, and let $\mathbf{Pa}_i$ be the set of parent nodes of $X_i$ in $S$. We let $x_i^k$ for $k = 1, \ldots, r_i$ denote the possible values taken by $X_i$ and let $\mathbf{pa}_i^j$ for $j = 1, \ldots, q_i$ denote the set of possible values taken by $\mathbf{Pa}_i$. (Where, as usual, a value for $\mathbf{Pa}_i$ determines a value for each of the variables in $\mathbf{Pa}_i$.) We can now define a parameter $\theta_{ijk}$ to represent the conditional probability table entry $P(X_i = x_i^k \mid \mathbf{Pa}_i = \mathbf{pa}_i^j)$. Finally, we use $\theta$ to denote the entire vector of parameter values $\theta_{ijk}$.

### 2.1    The basic equations

Our task is as follows. We have a current model (assignment of parameters) $\bar{\theta}$. We also have some set of (new or previously used) data cases $D = \{\mathbf{y}_1, \ldots, \mathbf{y}_N\}$. Each data case $\mathbf{y}_l$ is a (possibly) partial assignment of values to variables in the network. We want to construct a new model $\tilde{\theta}$ based on $\bar{\theta}$ and $D$.

One important metric for our choice of $\tilde{\theta}$ is the extent to which it explains our data $D$. We quantify that, as usual, via the log likelihood of $D$ in $\tilde{\theta}$, $\log P_{\tilde{\theta}}(D)$. However, the log-likelihood should not be the only factor. Since our current model $\bar{\theta}$ is already the result of some previous learning process (whether from the same sample set $D$ or from other samples), we do not want to completely ignore it, as we would if we based our choice of $\tilde{\theta}$ purely on the log-likelihood. Thus, we want to balance potential increases to the log-likelihood with the extent to which we move away from our current model.

We therefore choose $\tilde{\theta}$ so that it maximizes the following function $F$, which balances these two factors:

$$F(\tilde{\theta}) = \eta L_D(\tilde{\theta}) - d(\tilde{\theta}, \bar{\theta}). \qquad (1)$$

The first term, $L_D(\tilde{\theta})$, is the *normalized* log-likelihood of $D$ in the new model $\tilde{\theta}$, namely, $L_D(\tilde{\theta}) = \frac{1}{N} \sum_{l=1}^{N} \log P_{\tilde{\theta}}(\mathbf{y}_l)$. We use the normalized likelihood (rather than the un-normalized version) so as to eliminate the explicit dependency on the number of samples in our analysis. The "penalty" term $d(\tilde{\theta}, \bar{\theta})$ is an estimate of the distance between the new and old models. Its effect is to keep $\tilde{\theta}$ close to $\bar{\theta}$. The parameter $\eta > 0$ is a learning rate, which determines the extent to which we are willing to let our samples move us away from our current model.

Since it is computationally difficult to maximize (1), we choose to maximize a simplified version, obtained by linearizing the log-likelihood term. Let $\nabla L_D(\theta)$ be the gradient vector of $L_D(\theta)$. Let $\nabla_{ijk} L_D(\theta)$ be the entry in the gradient vector corresponding to the parameter $\theta_{ijk}$. Thus, in the vicinity of $\bar{\theta}$, we approximate the log-likelihood by its first order Taylor approximation,

$$L_D(\theta) \approx L_D(\bar{\theta}) + \nabla L_D(\bar{\theta}) \cdot (\theta - \bar{\theta}).$$

The above linear approximation degrades the further we move from the old parameter vector $\bar{\theta}$. However, by subtracting the distance term $d(\tilde{\theta}, \bar{\theta})$ from $L_D(\tilde{\theta})$, we can force the new parameter vector $\tilde{\theta}$ to stay relatively close to $\bar{\theta}$. Therefore, assuming that $\tilde{\theta}$ will not be too far away from $\bar{\theta}$, we can replace the $L_D(\tilde{\theta})$ term in (1), thereby changing our task to one of maximizing:

$$\hat{F}(\tilde{\theta}) = \eta [L_D(\bar{\theta}) + \nabla L_D(\bar{\theta}) \cdot (\tilde{\theta} - \bar{\theta})] - d(\tilde{\theta}, \bar{\theta}). \quad (2)$$

It turns out (see [2]) that the decomposition of the distribution $P_\theta$ implied by the network structure results in a particularly simple expression for the gradient vector:

$$\begin{aligned} \nabla_{ijk} L_D(\theta) &= \frac{1}{\theta_{ijk}} \frac{\sum_{l=1}^{N} P_\theta(x_i^k, \mathrm{pa}_i^j \mid \mathbf{y}_l)}{N} \\ &= \frac{\mathcal{E}_\theta(x_i^k, \mathrm{pa}_i^j \mid D)}{\theta_{ijk}}, \end{aligned} \quad (3)$$

where $\mathcal{E}_\theta(x_i^k, \mathrm{pa}_i^j \mid D)$ denotes $1/N \cdot \sum_{l=1}^{N} P_\theta(x_i^k, \mathrm{pa}_i^j \mid \mathbf{y}_l)$, i.e., the sample-based average of $x_i^k, \mathrm{pa}_i^j$.

The maximization problem in (2) is a constrained optimization problem, since our solution must be a legal assignment of CPT entries in the network. That is, for every $i, j$, we must have that $\sum_k \tilde{\theta}_{ijk} = 1$. Introducing Lagrange multipliers for these constraints, we conclude that our solution $\tilde{\theta}$ must satisfy $\frac{\partial}{\partial \tilde{\theta}_{ijk}} (\hat{F}(\tilde{\theta}) + \sum_{i'j'} \gamma_{i'j'} (\sum_{k'} \tilde{\theta}_{i'j'k'} - 1)) = 0$ for each $i, j, k$. That is,

$$\eta \nabla_{ijk} L_D(\bar{\theta}) - \frac{\partial}{\partial \tilde{\theta}_{ijk}} d(\tilde{\theta}, \bar{\theta}) + \gamma_{ij} = 0, \quad (4)$$

for all $i, j, k$. The result of solving this set of simultaneous equations will allow us to find the vector $\tilde{\theta}$ which maximizes (2), as desired. The answer, of course, depends on our choice of $d$. We now analyze this expression for three choices of $d$.

## 2.2 $\mathcal{L}_2$-norm

We define the $\mathcal{L}_2$-norm based distance between two parameter vectors $\tilde{\theta}$ and $\bar{\theta}$ to be:

$$\frac{1}{2} \|\tilde{\theta} - \bar{\theta}\|_2^2 = \frac{1}{2} \sum_{ijk} (\tilde{\theta}_{ijk} - \bar{\theta}_{ijk})^2.$$

Plugging this expression into (4), we get that

$$\eta \nabla_{ijk} L_D(\bar{\theta}) - (\tilde{\theta}_{ijk} - \bar{\theta}_{ijk}) + \gamma_{ij} = 0 \quad (5)$$

for all $i, j, k$.

Summing over $k$, i.e., the $r_i$ different possible values that $X_i$ can take, we get that:

$$\eta \sum_k \nabla_{ijk} L_D(\bar{\theta}) - (1 - 1) + r_i \gamma_{ij} = 0.$$

Thus, $\gamma_{ij} = \frac{\eta}{r_i} \sum_k \nabla_{ijk} L_D(\bar{\theta})$. If we plug these values of $\gamma_{ij}$ back into (5), we obtain the following update rule for $\tilde{\theta}$:

$$\tilde{\theta}_{ijk} = \bar{\theta}_{ijk} + \eta (\nabla_{ijk} L_D(\bar{\theta}) - \frac{1}{r_i} \sum_{k'} \nabla_{ijk'} L_D(\bar{\theta})). \quad (6)$$

If we substitute for $\nabla_{ijk} L_D(\bar{\theta})$ its value according to (3), we obtain precisely the standard gradient projection algorithm described in [2].

## 2.3 Relative entropy

The two parameter vectors, with the associated network structure, define two probability distributions over the the same space—the joint probability space of the variables in the Bayesian network. Thus, we can compare the distance between the two parameter vectors using the distance between the distributions they induce over this space.

One of the widely-used distance measures for two distributions over the same space is the *relative entropy* (also known as KL-divergence [14]). Given two distributions $p$ and $q$ over some joint probability space $\mathbf{X}$, the relative entropy between $p$ and $q$ is defined to be

$$d_{KL}(p \| q) = \sum_{\mathbf{x}} p(\mathbf{x}) \log \frac{p(\mathbf{x})}{q(\mathbf{x})}.$$

While this definition does not seem particularly amenable to analysis within (4), the relative entropy between two distributions over the same network has the following very nice decomposition property:

$$d_{KL}(\tilde{\theta} \| \bar{\theta}) = \sum_i \sum_j P_{\tilde{\theta}}(\mathbf{Pa}_i = \mathrm{pa}_i^j) \, d_{KL}(\tilde{\theta}_{ij} \| \bar{\theta}_{ij}); \quad (7)$$

where $\bar{\theta}_{ij}$ is the vector $(\tilde{\theta}_{ij1}, \ldots, \tilde{\theta}_{ijr_i})$, or (equivalently) the distribution over the different values $x_i^k$ of $X_i$ defined by $P_{\tilde{\theta}}(X_i \mid \mathbf{Pa}_i = \mathrm{pa}_i^j)$.

To take the derivative, we first replace $P_{\bar{\theta}}(\mathbf{Pa}_i = \mathbf{pa}_i^j)$ with some known estimate $\hat{P}(\mathbf{pa}_i^j)$ (whose exact nature will be determined later). It now follows that:

$$\frac{\partial}{\partial \tilde{\theta}_{ijk}} d_{KL}(\tilde{\theta}\|\bar{\theta}) = \hat{P}(\mathbf{pa}_i^j) \left( \log \frac{\tilde{\theta}_{ijk}}{\bar{\theta}_{ijk}} + 1 \right).$$

Plugging this value into (4), we get that:

$$\eta \nabla_{ijk} L_D(\bar{\theta}) - \hat{P}(\mathbf{pa}_i^j) \left( \log \frac{\tilde{\theta}_{ijk}}{\bar{\theta}_{ijk}} + 1 \right) + \gamma_{ij} = 0.$$

After some simple algebraic manipulation, and plugging in the value for the gradient as in (3), we get:

$$\tilde{\theta}_{ijk} = \frac{\bar{\theta}_{ijk} \exp\left(\frac{\eta}{\bar{\theta}_{ijk}} \frac{\mathcal{E}_{\bar{\theta}}(x_i^k, \mathbf{pa}_i^j | D)}{\hat{P}(\mathbf{pa}_i^j)}\right)}{\sum_k \bar{\theta}_{ijk} \exp\left(\frac{\eta}{\bar{\theta}_{ijk}} \frac{\mathcal{E}_{\bar{\theta}}(x_i^k, \mathbf{pa}_i^j | D)}{\hat{P}(\mathbf{pa}_i^j)}\right)}. \quad (8)$$

### 2.4  $\chi^2$ distance

The $\chi^2$ distance [5] between two distributions $p$ and $q$ as above is:

$$\chi^2(p\|q) = \frac{1}{2} \sum_{\mathbf{x}} (p(\mathbf{x}) - q(\mathbf{x}))^2 / q(\mathbf{x}).$$

This distance function is, in fact, a linear approximation of the relative entropy distance [5]. We use the $\chi^2$ distance to approximate the relative entropy by first decomposing the relative entropy as in (7), and then using the $\chi^2$ distance to approximate each of the distributions distributions $P_{\bar{\theta}}(X_i | \mathbf{Pa}_i = \mathbf{pa}_i^j)$ which appear in the summation. That is, we use as our distance function

$$\hat{\chi}^2(\tilde{\theta}\|\bar{\theta}) = \sum_i \sum_j P_{\bar{\theta}}(\mathbf{Pa}_i = \mathbf{pa}_i^j) \, \chi^2(\tilde{\theta}_{ij} \| \bar{\theta}_{ij}). \quad (9)$$

Approximating $P_{\bar{\theta}}(\mathbf{pa}_i^j)$ using $\hat{P}(\mathbf{pa}_i^j)$, as before, and then taking the derivative, we get $\frac{\partial}{\partial \tilde{\theta}_{ijk}} \hat{\chi}^2(\tilde{\theta}\|\bar{\theta}) = \hat{P}(\mathbf{pa}_i^j)(\tilde{\theta}_{ijk}/\bar{\theta}_{ijk} - 1)$. Plugging this result into (4), we get: $\eta \nabla_{ijk} L_D(\bar{\theta}) - \hat{P}(\mathbf{pa}_i^j)(\tilde{\theta}_{ijk}/\bar{\theta}_{ijk} - 1) + \gamma_{ij} = 0$. Thus,

$$\tilde{\theta}_{ijk} = \frac{\eta}{\hat{P}(\mathbf{pa}_i^j)} \nabla_{ijk} L_D(\bar{\theta}) \bar{\theta}_{ijk} + \frac{\gamma_{ij}}{\hat{P}(\mathbf{pa}_i^j)} \bar{\theta}_{ijk} + \bar{\theta}_{ijk}. \quad (10)$$

Substituting $\nabla_{ijk} L_D(\bar{\theta})$ for its value according to (3), and summing over $k$ to compute $\gamma_{ij}$, we get:

$$1 = \frac{\eta}{\hat{P}(\mathbf{pa}_i^j)} \sum_k \mathcal{E}_{\bar{\theta}}(x_i^k, \mathbf{pa}_i^j | D) + \frac{\gamma_{ij}}{\hat{P}(\mathbf{pa}_i^j)} + 1,$$

and therefore $\gamma_{ij} = -\eta \mathcal{E}_{\bar{\theta}}(\mathbf{pa}_i^j | D)$. Now, returning to (10), we have that

$$\tilde{\theta}_{ijk} = \frac{\eta}{\hat{P}(\mathbf{pa}_i^j)} \mathcal{E}_{\bar{\theta}}(x_i^k, \mathbf{pa}_i^j | D) - \quad (11)$$

$$\frac{\eta}{\hat{P}(\mathbf{pa}_i^j)} \mathcal{E}_{\bar{\theta}}(\mathbf{pa}_i^j | D) \bar{\theta}_{ijk} + \bar{\theta}_{ijk}.$$

## 3  Batch update rules

In the previous section, we derived some basic rules (corresponding to three different distance functions) for updating a given parameter vector $\bar{\theta}$ given some data set $D$. The interpretation of these rules is different in the batch and on-line contexts. We begin with the batch learning task, since it has received far more attention, so that the problem and the metrics for evaluating solutions are much better understood. Thus, we assume that we are given a fixed data set $D$, and that our goal is to find a parameter vector $\tilde{\theta}$ which best explains $D$. When the likelihood function cannot be computed analytically, as in the case of Bayesian networks, learning algorithms of this type usually employ a hill climbing scheme that requires multiple iterations, where in each iteration the current set of parameters is used to derive a new set of parameters.

This type of procedure fits directly into our the basic framework described in the previous section. Our current model $\bar{\theta}$ is the one resulting from the previous pass over $D$; the goal of the current pass is to update $\bar{\theta}$, resulting in a new model $\tilde{\theta}$. Therefore, we can instantiate the three update rules described above in the context of this problem.

We have already seen that the update rule for the $\mathcal{L}_2$-norm leads directly to the gradient projection algorithm of [3, 2]. In order to apply the other two update rules in this context, we must only decide on the estimate $\hat{P}(\mathbf{pa}_{ij})$, introduced as an approximation to $P_{\bar{\theta}}(\mathbf{Pa}_i = \mathbf{pa}_i^j)$. In a batch setting, a reasonable solution is to use the *sample-based expectation* as an approximation:

$$\hat{P}(\mathbf{pa}_{ij}) = \frac{1}{N} \sum_{l=1}^N P_{\bar{\theta}}(\mathbf{Pa}_i = \mathbf{pa}_i^j | \mathbf{y}_l) = \mathcal{E}_{\bar{\theta}}(\mathbf{pa}_i^j | D). \quad (12)$$

Let us begin by considering the $\chi^2$ update rule. Plugging in this last expression into (11), we get:

$$\tilde{\theta}_{ijk} = \frac{\eta \mathcal{E}_{\bar{\theta}}(x_i^k, \mathbf{pa}_i^j | D)}{\mathcal{E}_{\bar{\theta}}(\mathbf{pa}_i^j | D)} - \frac{\eta \mathcal{E}_{\bar{\theta}}(\mathbf{pa}_i^j | D)}{\mathcal{E}_{\bar{\theta}}(\mathbf{pa}_i^j | D)} \cdot \bar{\theta}_{ijk} + \bar{\theta}_{ijk}$$

$$= \eta \frac{\mathcal{E}_{\bar{\theta}}(x_i^k, \mathbf{pa}_i^j | D)}{\mathcal{E}_{\bar{\theta}}(\mathbf{pa}_i^j | D)} + (1 - \eta) \bar{\theta}_{ijk}. \quad (13)$$

This equation describes a weighted average between the the parameter obtained by dividing the sample-based average of the pair $x_i^k, pa_i^j$ by those for $\mathbf{pa}_i^j$ and the current parameter $\bar{\theta}_{ijk}$. When $\eta = 1$, this update rule reduces to the standard EM algorithm. We therefore call this parameterized update rule EM($\eta$).

For $\eta < 1$, EM($\eta$) instantiates the new parameter values to be a weighted combination between the EM update and the current vector of parameters. (A form of parameter update that belongs to the family of *stochastic approximation* algorithms [20, 7].) The new parameters are therefore somewhere between the old ones and the ones induced by the data. Thus, EM($\eta$) updates the parameter values more slowly than standard EM.

For $\eta > 1$, EM($\eta$) does the reverse: rather than interpolating between these two points, it uses the parameters induced from the data to extrapolate the direction of update. It then speeds up the update process, going even further in that direction than what is implied by the data. While this extrapolation might seem somewhat counterintuitive, it has been used successfully in density estimation problems [18, 19]. As we show in the next two sections, this faster update rate can speed up convergence considerably.

Finally, we consider the relative entropy update rule (8). Using (12) as our estimate for $\hat{P}(\mathbf{pa}_i^j)$, we get:

$$\tilde{\theta}_{ijk} = \frac{\bar{\theta}_{ijk} \exp(\frac{\eta}{\bar{\theta}_{ijk}} \frac{\mathcal{E}_{\bar{\theta}}(x_i^k, \mathbf{pa}_i^j | D)}{\mathcal{E}_{\bar{\theta}}(\mathbf{pa}_i^j | D)})}{\sum_k \bar{\theta}_{ijk} \exp(\frac{\eta}{\bar{\theta}_{ijk}} \frac{\mathcal{E}_{\bar{\theta}}(x_i^k, \mathbf{pa}_i^j | D)}{\mathcal{E}_{\bar{\theta}}(\mathbf{pa}_i^j | D)})} . \quad (14)$$

Essentially, each parameter $\bar{\theta}_{ijk}$ is multiplied by a factor which is exponential in the relative difference between the parameters induced by the data (via the ratio of sample based expectations) and the current values of the parameters. Intuitively, larger differences cause the parameter to be updated more rapidly. Again, $\eta$ serves to guide the rate at which the parameters are changed. This update rule is called EG($\eta$) (see [13, 11]), due to its use of an exponentiated gradient as its main term. We note that, in our experimental results, the batch version of EG($\eta$) performed quite poorly. Therefore, we focus the rest of our discussion of batch update on the EM($\eta$) procedure.

## 4 Convergence properties

Since the likelihood function of Bayesian networks with missing data has multiple local maxima, it is impossible to derive global convergence bounds on the performance of the any local parameter update scheme. We therefore show in this section results about the uniform rate of *local* convergence of the EM($\eta$)-update rule. More precisely, we analyze the rate of convergence of EM($\eta$) in a neighborhood of some local maximum $\theta^\star$. Throughout this section we assume that $\theta^\star$ is in the interior of the simplex of feasible parameters, that is, $\forall i, j, k : 1 > \theta_{ijk}^\star > 0$. (Similar convergence results can be obtained when the solution is on the simplex boundary.)

The basic technique is as follows. We view the EM($\eta$) update rule as an operator over parameter vectors $\theta$. In a sufficiently small neighborhood of $\theta^\star$, we can approximate this operator using the linear component of its Taylor expansion. The convergence rate of a linear operator is determined by the eigen values of the matrix which describes it. By analyzing these eigen values, we get precise bounds on the rate of convergence in the vicinity of $\theta^\star$.

We begin by defining EM($\eta$) as an operator $\Phi$. Formally,

$$\Phi(\theta)_{ijk} = \eta \frac{\sum_l P_\theta(x_i^k, \mathbf{pa}_i^j | \mathbf{y}_l)}{\sum_{k',l} P_\theta(x_i^{k'}, \mathbf{pa}_i^j | \mathbf{y}_l)} + (1-\eta)\theta_{ijk} . \quad (15)$$

We want to analyze the behavior of $\Phi$ around one of the local maxima $\theta^\star$. Our first observation is that $\theta^\star$ is necessarily a fixpoint of $\Phi$.

**Lemma 1:** If $\theta^\star$ is a local maximum of the likelihood function, then $\Phi(\theta^\star) = \theta^\star$, for any value of $\eta$.

**Proof:** Due to the constraint $\sum_k \theta_{ijk} = 1$, we get (by introducing the appropriate Lagrange multiplier) that the following equation must hold at any local maximum:

$$\frac{\partial}{\partial \theta_{ijk}} \left[ L_D(\theta) + \gamma(\sum_{k'} \theta_{ijk'} - 1) \right]_{\theta=\theta^\star} = 0 .$$

Therefore,

$$\frac{1}{N\theta_{ijk}^\star} \sum_{l=1}^N P_{\theta^\star}(x_i^k, \mathbf{pa}_i^j | \mathbf{y}_l) + \gamma = 0 . \quad (16)$$

Multiplying the equation by $\theta_{ijk}^\star$ and summing over all $k$ we get, $1/N \sum_{k,l} P_{\theta^\star}(x_i^k, \mathbf{pa}_i^j | \mathbf{y}_l) + \gamma \sum_k \theta_{ijk}^\star = 0$ and therefore that $\gamma = -\frac{1}{N} \sum_{k',l} P_{\theta^\star}(x_i^{k'}, \mathbf{pa}_i^j | \mathbf{y}_l)$. Substituting this value for $\gamma$ in (16), we get

$$\frac{1}{N\theta_{ijk}^\star} \sum_l P_{\theta^\star}(x_i^k, \mathbf{pa}_i^j | \mathbf{y}_l) = \frac{1}{N} \sum_{k',l} P_{\theta^\star}(x_i^{k'}, \mathbf{pa}_i^j | \mathbf{y}_l) .$$

and finally that

$$\theta_{ijk}^\star = \frac{\sum_l P_{\theta^\star}(x_i^k, \mathbf{pa}_i^j | \mathbf{y}_l)}{\sum_{k',l} P_{\theta^\star}(x_i^{k'}, \mathbf{pa}_i^j | \mathbf{y}_l)} , \quad (17)$$

which immediately implies that $\theta^\star = \Phi(\theta^\star)$, as desired. ∎

Letting $\tilde{\theta}$ denote $\Phi(\theta)$, we now have:

$$\tilde{\theta} - \theta^\star = \Phi(\theta) - \Phi(\theta^\star) = \nabla \Phi(\theta^\star)(\theta - \theta^\star) + o(\theta - \theta^\star) . \quad (18)$$

The term $\nabla \Phi(\theta^\star)$ is an $m \times m$ matrix (where $m$ is the dimension of $\theta$), whose $uv$th entry is the derivative of the $u$th component of $\Phi$ with respect to the $v$th parameter, evaluated at the local maximum likelihood parameter set $\theta^\star$.

In the vicinity of $\theta^\star$, $\nabla \Phi(\theta^\star)(\theta - \theta^\star)$ forms a linear approximation to $\Phi$. We therefore analyze this derivative, and show that, in the vicinity of $\theta^\star$, for $0 < \eta < 2$, there exists a norm $\| \cdot \|$ on $\mathbb{R}^m$ and a number $0 \leq \rho_\eta < 1$ such that

$$\|\nabla \Phi(\theta^\star)(\theta - \theta^\star)\| \leq \rho_\eta \|\theta - \theta^\star\| .$$

Thus, the operator $\Phi$ forms a contraction mapping around $\theta^\star$, one which induces a convergence rate of $\lambda$.

Our analysis generalizes a technique first used by Peters and Walker [18] in the context of estimating the parameters of a mixture of normal distributions. For simplicity, assume that all the parameters of the network are known and fixed, except for $\theta_{ijk'}$ for $k' = 1, \ldots, r_i$. (The general case is sketched at the end of the section.) Hence, we can take $m = r_i$.

Let $\alpha_{ijk}^l$ denote $P_{\theta^\star}(x_i^k, \mathbf{pa}_i^j \mid \mathbf{y}_l)$. Let $\delta_{kk'} = 1$ if $k = k'$ and 0 otherwise. Then, by a process of taking derivatives and algebraic simplifications, we get:

$$\nabla \Phi_{kk'} = \frac{\partial \Phi(\theta)_{ijk}}{\partial \theta_{ijk'}} = \delta_{kk'}(1-\eta) +$$
$$\frac{\eta}{\theta_{ijk'} \sum_{k''l} \alpha_{ijk''}^l} \left( \delta_{kk'} \sum_l \alpha_{ijk}^l - \sum_l \alpha_{ijk}^l \alpha_{ijk'}^l \right) .$$

Since at the local maximum $\theta^\star$, we have that $\theta_{ijk}^\star = \sum_l \alpha_{ijk}^l / \sum_{k'',l} \alpha_{ijk''}^l$ (Equation (17)), then

$$\nabla \Phi_{kk'} = \delta_{kk'} - \eta \frac{\sum_l \alpha_{ijk}^l \alpha_{ijk'}^l}{\theta_{ijk'} \sum_{k''l} \alpha_{ijk''}^l} ,$$

or in a matrix form,

$$\nabla \Phi(\theta^\star) = I - \eta M \text{ where } M_{kk'} = \frac{\sum_l \alpha_{ijk}^l \alpha_{ijk'}^l}{\theta_{ijk'} \sum_{k''l} \alpha_{ijk''}^l} .$$

The matrix $M$ can be decomposed into two matrices $M = QR$ where $Q$ is a diagonal matrix with $Q_{kk} = 1/\theta_{ijk}$, and $R$ is an $m \times m$ matrix with $R_{kk'} = \frac{\sum_l \alpha_{ijk}^l \alpha_{ijk'}^l}{\sum_{k''l} \alpha_{ijk''}^l}$. Denote by $\vec{\alpha}_l$ the vector $(\alpha_{ij1}^l, \alpha_{ij2}^l, \ldots, \alpha_{ijm}^l)$, and let $\beta = \sum_{kl} \alpha_{ijk}^l$. The matrix $R$ can be rewritten as $R = \frac{1}{\beta} \sum_l \vec{\alpha}_l^T \vec{\alpha}_l$. Clearly, $Q$ is symmetric and positive definite if $\theta_{ijk} > 0$. $R$ is symmetric and positive semi-definite since for any $\xi \in \mathbb{R}^m$,

$$\xi R \xi^T = \frac{1}{\beta} \sum_l \xi \left( \vec{\alpha}_l^T \vec{\alpha}_l \right) \xi^T = \frac{1}{\beta} \sum_l \left( \vec{\alpha}_l \xi^T \right)^T \left( \vec{\alpha}_l \xi^T \right)$$
$$= \frac{1}{\beta} (\xi \cdot \vec{\alpha}_l)^2 \geq 0 .$$

We can now define the norm for which $\Phi$ is a contraction. Intuitively, $Q$ is a diagonal matrix, and therefore has no influence on the rate of contraction. The rate of contraction is determined by $R$. Hence, our norm should factor out $Q$, so as to isolate the influence of $R$. Therefore, for a vector $\xi$, we define $\|\xi\|$ to be $\xi Q^{-1} \xi^T$.

As we discuss below, the rate of contraction with respect to this norm is essentially determined by the eigen values of $R$. We therefore begin by analyzing these eigen values. We first show that the eigen values of $R$ all lie in $[0, 1]$. We use the following lemmas [12]:

**Lemma 2:** Let $A$ and $B$ be two matrices of dimensions $n \times m$ and $m \times n$, respectively, where $m \geq n$. Then, every eigen value of $AB$ is also an eigen value of $BA$.

**Lemma 3:** Denote by $\lambda_{max}(A)$ the largest eigen value of a matrix $A$. Let $\{A_i\}_{i=1}^N$ be a set of $N$ matrices of the same dimension. Then, $\lambda_{max}\left(\sum_{i=1}^N A_i\right) \leq \sum_{i=1}^N \lambda_{max}(A_i)$.

Based on these lemmas, we get

$$\begin{aligned}
\lambda_{max}(R) &= \frac{1}{\beta} \lambda_{max}\left(\sum_{l=1}^N \vec{\alpha}_l^T \vec{\alpha}_l\right) \\
&\leq \frac{1}{\beta} \sum_{l=1}^N \lambda_{max}(\vec{\alpha}_l^T \vec{\alpha}_l) = \frac{1}{\beta} \sum_{l=1}^N \lambda_{max}(\vec{\alpha}_l \vec{\alpha}_l^T) \\
&= \frac{1}{\beta} \sum_{l=1}^N \|\vec{\alpha}_l\|_2^2 \leq \frac{1}{\beta} \sum_{l=1}^N \|\vec{\alpha}_l\|_1 \\
&= \frac{1}{\beta} \sum_{l=1}^N \sum_k \alpha_{ijk}^l = \frac{\beta}{\beta} = 1 .
\end{aligned}$$

Now, let $\lambda_1, \ldots, \lambda_p$ be the non-zero eigen values of $R$, and let $\Omega$ be the subspace defined by the corresponding eigen vectors $\mathbf{v}_1, \ldots, \mathbf{v}_p$. We begin by considering the convergence properties of $\Phi$ within $\Omega$. More precisely, assume that $\theta$ and $\theta^\star$ are both within $\Omega$. Now, consider $\xi = \theta - \theta^\star$. Since $\xi$ is also in $\Omega$, it is expressible as a linear combination of the $\mathbf{v}_i$'s. The application of $R$ to $\xi$ causes the linear coefficient of each $\mathbf{v}_i$ to be multiplied by a factor of $\lambda_i$. Thus, the rate at which the various components of $\xi$ shrink as $R$ is applied is determined by the size of the corresponding eigen values.

Let $\lambda_{min}$ and $\lambda_{max}$ be the smallest and largest non-zero eigen value of $R$, respectively. The rate at which the operator $\nabla \Phi(\theta^\star) = I - \eta QR$ contracts its various components is determined by its largest eigen value. It is easy to see that this value is the larger of $|1 - \eta \lambda_{min}|$ and $|1 - \eta \lambda_{max}|$. Formally, the *operator norm* of $\nabla \Phi(\theta^\star)$ within $\Omega$ and with respect to the above norm is $\rho_\eta = \max\{|1 - \eta \lambda_{min}|, |1 - \eta \lambda_{max}|\}$. For $0 < \eta < 2$, we get that $\rho_\eta < 1$, so that

$$\|\nabla \Phi(\theta^\star)(\theta - \theta^\star)\| \leq \rho_{eta} \|\theta - \theta^\star\| < \|\theta - \theta^\star\| . \quad (19)$$

Thus, for any $\eta$ in the range 0 to 2, we are guaranteed convergence of $\text{EM}(\eta)$ to $\theta^\star$ in a neighborhood around it.

Note that the above argument only applies to vectors that are within the subspace $\Omega$. The vector $\theta^\star$ in this subspace, since it is a linear combination of the $\vec{\alpha}_l$, but $\theta$ may not be. However, since $\Phi(\theta)$ is a weighted average of $\theta$ and a vector in $\Omega$ (a weighted average of the $\vec{\alpha}_l$), the component of $\theta$ which is not in $\Omega$ shrinks by a factor of $|1 - \eta|$ with every application of $\Phi$. Therefore, for $0 < \eta < 2$, we have that repeated applications of $\Phi$ to a vector $\theta$ result in exponential convergence to this subspace.

The more general case when all the variables of $\theta$ are updated uses a similar, albeit more complex, derivation. Roughly speaking, in the general case, the matrix $M$ is a block matrix where the $i$th block is an $(r_i q_i) \times (r_i q_i)$ square matrix. We use the fact that $\sum_{jk} \alpha_{ijk}^l = \sum_{jk} P(x_i^k, \mathbf{pa}_i^j \mid \mathbf{y}_l) = 1$ to show that each block is a positive semi-definite matrix with eigen values smaller than 1 and use this to bound the entire operator norm, as above. We therefore obtain the following theorem:

**Theorem 1:** For any data set $D$, and any $0 < \eta < 2$, the $\text{EM}(\eta)$ update rule (Equation (13)) converges to a local maximum of the likelihood function $P_{\theta^\star}(D)$, within some neighborhood $\|\theta - \theta^\star\| < \delta$.

While local convergence to a local maximum is guaranteed for any $\eta$ in the right range, there is a particular value $\eta^\star$, called "the optimal learning rate", which yields, within a neighborhood of a local maximum $\theta^\star$, the fastest uniform rate of convergence of the $\text{EM}(\eta)$ update rule. From the derivation above, the optimal rate of convergence is obtained when the contraction factors, $|1 - \eta \lambda_{min}|$ and $|1 - \eta \lambda_{max}|$ are equal. Since $0 < \lambda_{min} \leq \lambda_{max} \leq 1$ this equality is obtained when $1 - \eta^\star \lambda_{min} = \eta^\star \lambda_{max} - 1$. Hence,

$$\eta^\star = 2/(\lambda_{min} + \lambda_{max}) \geq 2/(1 + \lambda_{min}) > 1 .$$

Therefore, the local optimal rate of convergence is greater than 1, which implies that standard EM is inferior (in a neighborhood of the local maximum) to the EM($\eta^*$) update rule! Furthermore, if $\lambda_{max}$ is strictly smaller than one, the optimal learning rate $\eta^*$ is actually greater than 2, even though local convergence of the EM($\eta$)-update cannot be guaranteed for such value of $\eta$. In fact, we have observed in practice cases where an aggressive learning of $\eta > 2$ is beneficial.

The impact of the choice of $\eta$ is clear from (19). Each iteration of EM($\eta$) shrinks the distance between our current parameter vector $\theta$ to $\theta^*$ by a factor $\rho_\eta$. Thus, if $\rho_\eta < \rho_{\eta'}$, then the parameter vector $\theta$ produced by EM($\eta$) after some number $K$ of iterations will be closer to $\theta^*$ than the vector $\theta'$ produced by a similar number of iterations of EM($\eta'$). In fact, the distance between $\theta$ and $\theta^*$ will be smaller than the comparable distance for $\theta'$ by an exponential factor $(\rho_\eta/\rho_{\eta'})^K$!

Thus, substantial improvements in convergence can be obtained by running EM with the optimal learning rate $\eta^*$. Unfortunately, direct calculation of $\eta^*$ requires knowledge of $\lambda_{min}$ and $\lambda_{max}$ at the local maximum of interest. This prevents determination of $\eta^*$ prior to parameter estimation. However, if we assume that our current parameter vector $\theta$ is fairly close to $\theta^*$, then $\lambda_{min}$ and $\lambda_{max}$ at the current position can be used to provide a fairly good estimate for $\eta^*$. These eigenvalues can be estimated fairly efficiently using techniques similar to those used in [16]. Of course, empirical testing would be necessary to check whether the additional computational cost of approximating $\eta^*$ is worthwhile in practice.

Finally, we note that our convergence theorem and the result on the optimal rate of convergence hold only in a neighborhood of the local maximum. By contrast, the standard EM update rule is guaranteed to converge to a local maximum from any point in the space (except perhaps for pathological cases [6]). A similar guarantee cannot be made for other values of $\eta$. However, our experimental results in the next section show that, in practice, we do get convergence from random starting positions for $\eta > 1$.

We also note that the convergence rate analysis for EM($\eta$) was only for vectors within the subspace $\Omega$. The components of $\theta$ that are orthogonal to $\Omega$ contract with a rate of $|1 - \eta|$, which is clearly optimal when $\eta = 1$. By interspersing iterations of EM(1) with EM($\eta$) for large $\eta$, we can trade off convergence within $\Omega$ with convergence outside of $\Omega$. In practice, however, this has not turned out to be necessary. Furthermore, in the full paper we will show that, if our network structure allows a sufficient variety of qualitatively different data cases, and if our data set $D$ is sufficiently large, the vectors $\bar{\alpha}_i$ will (with high probability) span the entire space, in which case $R$ is guaranteed to have full dimension. Then, any vector $\theta$ is within the space $\Omega$, and our convergence rate argument applies.

## 5 Experimental results

We tested the performance of the EM($\eta$) algorithm in practice by using it for parameter estimation in two well-known networks: the Alarm network for ICU ventilator management [1] and the Insurance network for car insurance underwriting [2]. Since the results on the two networks were fairly comparable, we present major results for Alarm. Both individual runs of the EM($\eta$) update rule and 10-fold cross-validated update runs gave similar results; we present data from individual runs to show performance of the update rule over a single parameter estimation task.

To test the algorithm, we generated random training and test cases from the original network, made both types of data cases partially observable (as described below), estimated the network parameters from the training data, and then tested the performance of the resulting network on the test data. We attempted to introduce partial observability in a way that is compatible with the use of the network in practice. Thus, we partitioned the network variables into three categories: input nodes, hidden nodes, and output nodes. The hidden nodes correspond to variables like "Heart Rate", which are never observed in either the training or the test data. The input nodes are variables like "Heart Rate EKG", which are often observable both in the training data and when the network is used in practice. The output nodes are variables like "Left Ventricular Failure", whose value we are interested in querying. We therefore generated training data as follows: complete data cases were generated from the distribution in the network. In each data case, the values of the hidden variables were obscured; for each input and output variable, its value was obscured with some fixed probability.[1]

In the learning phase, we began with an initial random choice of network parameters. We proceeded to update parameters using an initial EM(1) iteration followed by a fixed number of EM($\eta$) iterations. For each task, we experimented with several values of $\eta$. We also experimented with various proportions of randomly unobserved variables in the training input observations. In each update run, we measured (i) average log marginal likelihood over non-hidden nodes, and (ii) absolute and relative error for the conditional probability of various output nodes given the (partial) instantiation of the input nodes in the training case.[2] We measured these parameters over both our training and test data.[3]

---

[1] Since the process of obscuring values does not depend on the actual values of any of the variables in the data case, this process is *ignorable* as defined by Rubin [21].

[2] If the probability of the relevant output node given the observed input nodes is $p$ in the learned network and $p^*$ in the correct network, then the absolute error is $|p-p^*|$ and the relative error is $(|p-p^*|)/p^*$. The distribution of missing data in the test cases was identical, in each case, to the distribution used in the training data.

[3] For comparison, we also experimented with training data where there were no hidden variables, and the missing values were simply selected uniformly over all variables in the network. The results were qualitatively similar to the ones shown here, and were omitted for lack of space.

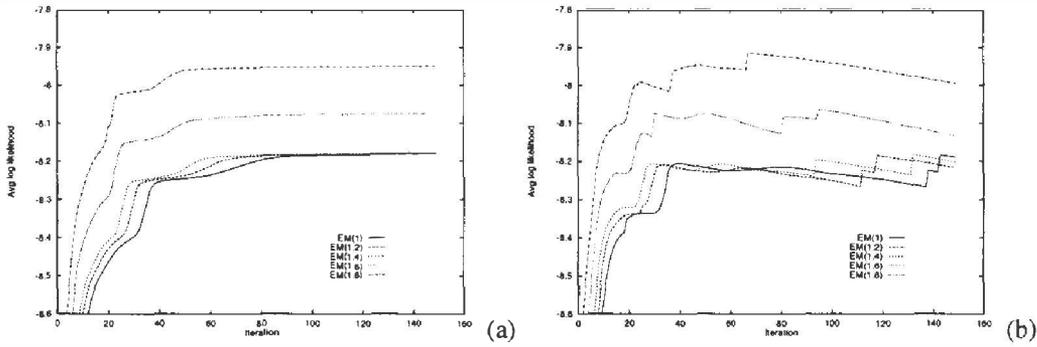

Figure 1: Average marginal log-likelihood for the Alarm network with 2000 training cases (a) over the training data, and (b) over 2000 test cases, using 20% unknown input values in both.

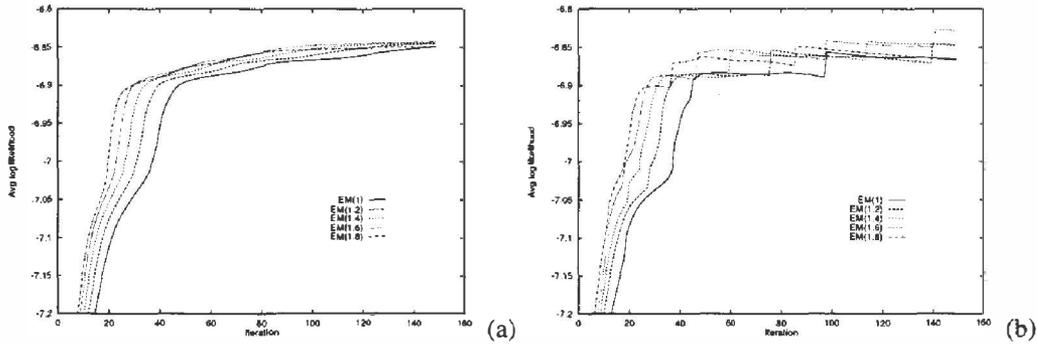

Figure 2: Average log marginal likelihood for the Alarm network with 2000 training cases (a) over the training data, and (b) over 2000 test cases, using 40% unknown input values in both. Here all values of $\eta$ converge to similar values.

Figures 1 and 2 show average log-likelihood convergence results with two proportions of unknowns in the input. The graphs are fairly typical, in that there is an initial phase of rapid convergence, with additional steep but smaller improvements later on. Between the phases where significant progress is made, there are long stretches where the log-likelihood improves only slightly. In these stretches, the log-likelihood of the training data continues ot increase consistently (as it should). However, the log-likelihood of the test data often decreases in these long stretches. This phenomenon is a typical example of overfitting the training data.

In both Figure 1 and 2, we see that the convergence rate of EM($\eta$) increases consistently with $\eta$. For high values of $\eta$, the difference in convergence rates can be quite dramatic. The initial convergence for EM(1.8) occurs around the 20th iteration, while EM(1) requires around 40 iterations. The difference becomes even more pronounced as the process continues, with EM(1.8) reaching a certain convergence level as much as 60 iterations before EM(1). This increasing separation between the different algorithms as the number of iterations grows supports our analysis of the impact of the convergence rate from the previous section.

The convergence in log-likelihood is reflected by a similar convergence in both error rates and parameter values, as shown in Figures 3 and 4. In all cases, we observe faster convergence using high values of $\eta$. In general, there is (as expected) a close correlation between the graphs for the different metrics—training set log-likelihood, test set log-likelihood, absolute and relative errors, and values for various parameters—e.g., in the iterations where significant progress takes place).

More interestingly, there is often noticeable repetition in convergence behavior over EM($\eta$) update runs for different values of $\eta$. This repetition or "stuttering" indicates that these runs follow a similar trajectory in parameter space, but that this process is simply much faster for larger values of $\eta$. This intuition is supported by the parameter graphs of Figure 4, which are typical of the graphs for other parameters in the network. Thus, higher values of $\eta$ accelerate the progress of the update algorithm, but the shape of the path within the parameter space is often preserved. This phenomenon raises interesting conjectures about the shape of the likelihood function over the parameter space.

However, the trajectories are not always identical. In Figure 1, EM(1.8) and EM(1.6) converge to a different log-likelihood value than EM with other values of $\eta$; likewise, in Figure 4(c), EM with higher values of $\eta$ converges to different parameters. In this case, it appears that the larger step size resulting from higher values of $\eta$ caused the parameter vector to move to a different region in the parameter space, resulting in convergence to a a different local maximum. In general, the exact local maximum to which EM($\eta$) converges depends both on $\eta$ and on the initial random assignment of parameter values. In this case, the higher values of $\eta$ converge to a better maximum in likelihood space, but this is not necessarily the case.

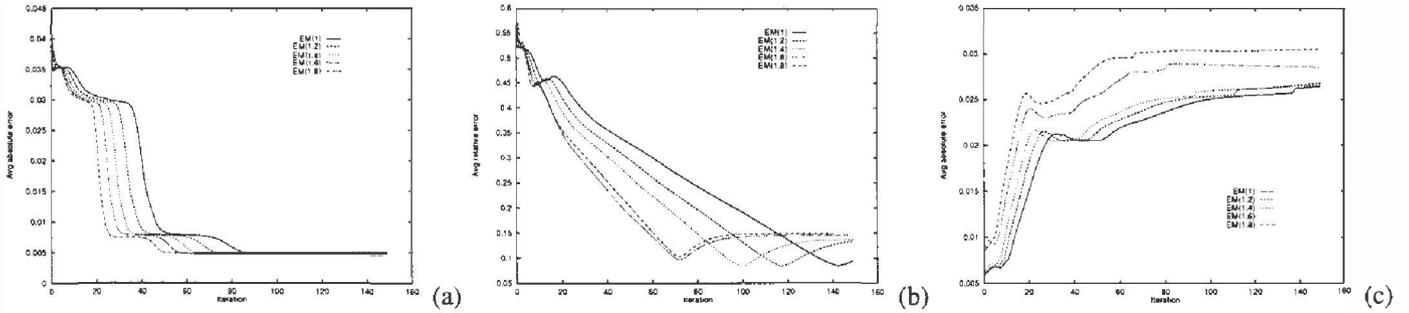

Figure 3: Error graphs for specific output nodes in the Alarm network over test cases. (a) Absolute error for "Left Ventricular Failure" node with 40% unknowns in the inputs; (b) Relative error for "Pulmonary Embolus" node with 20% unknowns in the inputs; (c) Absolute error for "Insufficient Anesthesia" node with 20% unknowns in the inputs.

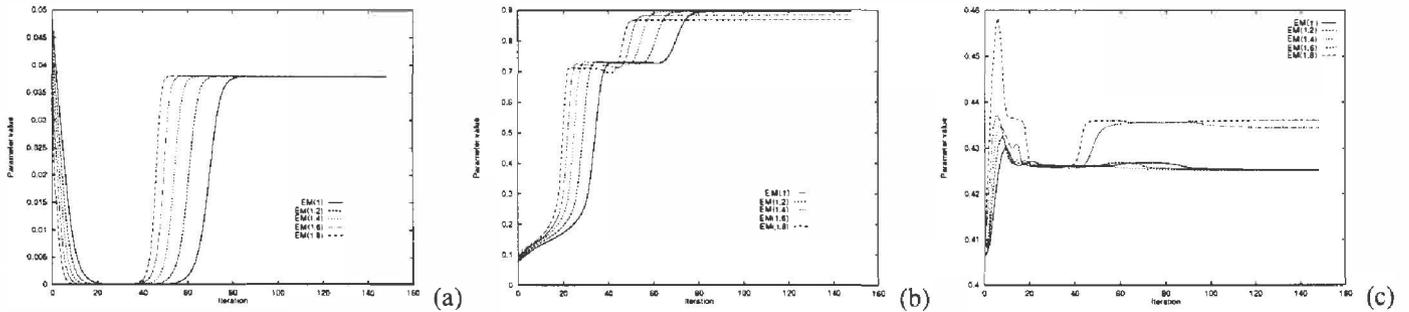

Figure 4: Conditional probability table value graphs for the Alarm network with 2000 training cases and 20% unknown input values for subparameters of (a) "Central Venous Pressure" node, (b) "Pulmonary Capillary Wedge Pressure" node, and (c) "Minute Volume" node.

We have also observed cases where EM(1.8) converges to a less optimal point. Furthermore, as we can see in Figure 3(b) and (c), a higher likelihood setting of parameters does not imply a lower error rate for all of the output variables of interest. A different parameter setting may improve the accuracy of some probability estimates while reducing the accuracy of others. In general, we can say nothing about the relative quality of the local maxima to which the different processes will converge or how they will affect the error for some specific output node. However, in all our experiments, higher values of $\eta$ resulted in faster convergence to whatever maximum the process ended up at.

As mentioned above, we also experimented with the EG($\eta$) update rule, with disappointing results. Using the EG($\eta$) rule often led to unstable updating, which made it difficult to converge upon a good set of parameters. Initial exploration of EG($\eta$)'s performance indicated a tendency to overcompensate for differences between estimated parameters and data. Such overcompensation caused parameters to quickly move away from good local maxima, which in turn caused other parameters to shift. The end result was often slow convergence to mediocre parameters. Despite this, more exploration of the EG($\eta$) algorithm is necessary to determine its effectiveness as a parameter estimation algorithm.

## 6 On-line parameter estimation

In an on-line setting, a learner observes one example at a time and needs to update its current model of the world. This task is very relevant in the context of Bayesian networks: An operational Bayesian network system is constantly presented with new cases for inference purposes. We would like the system to fine-tune itself based on these cases, adapting its network model to the environment. Nevertheless, the problem of on-line learning has been left largely unexplored in the context of Bayesian networks (with the exception of [17] and the recent work of [8] on structure learning with fully observable data). In this section, we take a first step towards providing a formal model for this task.

The key assumption in the on-line setting is that the learner cannot store past examples. Thus, one cannot simply accumulate lots of examples and then apply batch learning tasks. Learning takes place only by modifying a single hypothesis (or perhaps a few) that the learner maintains. Thus, the learner faces contradictory demands: it has to keep track of what has been learned so far while adjusting its hypothesis based on the new examples. The on-line learning framework of Kivinen and Warmuth [13] provides an analytical tool to balance these two requirements. In Section 2, we laid the foundation for applying these tools in the context of Bayesian networks. In this section, we interpret the resulting rules in the on-line setting.

In the on-line setting, the learner repeatedly gets one new sample $\mathbf{y}_t$ at a time. At each time step $t$, the learner has a current model, which we denote by $\theta^t$. An on-line update rule tells the learner how to transform $\theta^t$ into a new model $\theta^{t+1}$ based on the sample $\mathbf{y}_t$.

The basic framework is exactly as described in Section 2. At each step $t$, we want to solve Equation (1), with $D$ being our current sample $\mathbf{y}_t$. The gradient vector used in our approximation (2) is now the *instantaneous* gradient: $\nabla L_t(\boldsymbol{\theta}) = P_{\boldsymbol{\theta}}(x_i^k, \mathbf{pa}_i^j \mid \mathbf{y}_t)/\theta_{ijk}$.

The gradient projection update rule (6) remains unchanged. To instantiate our other update rules (8) and (11), we must define our approximation $\hat{P}$ for $P_{\bar{\boldsymbol{\theta}}}(\mathbf{pa}_i^j)$. Since we cannot store the previous examples, we simply use our current model $\boldsymbol{\theta}^t$ as the estimate for $\bar{\boldsymbol{\theta}}$. Thus, we set $\hat{P}(\mathbf{pa}_i^j) = P_{\boldsymbol{\theta}^t}(\mathbf{Pa}_i = \mathbf{pa}_i^j)$. Now, if we follow the same steps as in the batch setting with the above modifications we get that the EM($\eta$) and EG($\eta$) updates are,

$$\text{EM}(\eta): \quad \theta_{ijk}^{t+1} = \eta \frac{P_{\boldsymbol{\theta}^t}(x_i^k, \mathbf{pa}_i^j \mid \mathbf{y}_t)}{P_{\boldsymbol{\theta}^t}(\mathbf{pa}_i^j)} + (1-\eta)\theta_{ijk}^t,$$

$$\text{EG}(\eta): \quad \theta_{ijk}^{t+1} = \frac{1}{Z_{ij}^t} \theta_{ijk}^t \exp\left(\frac{\eta P_{\boldsymbol{\theta}^t}(x_i^k, \mathbf{pa}_i^j \mid \mathbf{y}_t)}{\theta_{ijk}^t P_{\boldsymbol{\theta}^t}(\mathbf{pa}_i^j)}\right).$$

where $Z_{ij}^t$ is a normalization constant for time step $t$.

As written, these update rules implicitly assume that the learning rate $\eta$ is the same for all $t$. However, as $t$ grows, our model $\boldsymbol{\theta}^t$ is based on more and more data cases. Intuitively, it seems that a new sample should have less effect on a well-established model (one based on many prior samples) than on a new one. Therefore, we may want to adapt our learning rate $\eta$ over time, based on the number of examples seen so far (see, for instance, [4]). This is, in fact, precisely the behavior we would get from a full Bayesian updating scheme for our model (a scheme which is unfortunately infeasible in the presence of partially-observable data cases [9]).

On the other hand, most of the worst case analyses of on-line learning algorithms do, in fact, employ a fixed learning rate to derive bounds on the performance of the on-line algorithm (see for instance [13] and the references therein). Furthermore, a fixed learning rate yielded very good results for other learning tasks, even when applied to natural data [10]. It is therefore an interesting and challenging research problem to determine whether an adaptive learning rate or a fixed one should be employed in on-line learning of Bayesian networks.

## 7 Conclusion and future work

In this paper, we re-examined the problem of parameter estimation in Bayesian networks. By applying a recently developed theoretical framework for this task [13], we derive two families of update rules EM($\eta$) and EG($\eta$), where EM(1) is simply the standard EM algorithm for Bayesian networks.

We applied the EM($\eta$) algorithm to the traditional batch learning task, and showed, both theoretically and empirically, that EM($\eta$) for values of $\eta$ larger than 1, converges to a locally maximal parameter assignment in fewer iterations than EM. Theoretically, we showed that for any (local or global) maximum of the likelihood function, there is a value $\eta^* > 1$ and a neighborhood around the local maximum, such that EM($\eta^*$) provides the fastest convergence (of any EM($\eta$) algorithm) to the maximum in that neighborhood.

We note that our theoretical convergence guarantees for EM($\eta$) are weaker than those for EM. In particular, EM is guaranteed to converge to a local maximum no matter its initial starting point. EM($\eta$), on the other hand, is only guaranteed to converge to any local maximum if its initial starting point is in a neighborhood of that maximum. However, our empirical results demonstrate that, in practice, this is not a concern. In almost all cases, EM($\eta$) does converge to a local maximum from a random initial position. In many cases, EM($\eta$) converges to the same local maximum as EM. In those cases where it converges to a different maximum, the outcome may be better or worse.

In all cases, however, the convergence of EM($\eta$) for large values of $\eta$ is significantly faster. It sometimes takes as many as 60 fewer iterations to reach the same point. The savings resulting from the accelerated convergence can be substantial: each iteration of EM involves running a Bayesian network inference algorithm on each one of the data cases in our training set. For large networks, a large number of data cases may be required in order to guarantee the robustness of our solution, and each application of Bayesian network inference can be very expensive. Moreover, the complexity of larger networks requires more iterations of EM to reach convergence. Since, as we discussed, the improvement of EM($\eta$) over EM grows exponentially with the number of iterations, the benefits of EM($\eta$) should be particularly significant in this context. Finally, it is important to note the main advantage of EM($\eta$) over other possible schemes for accelerating EM: EM($\eta$) is no more complicated to implement than standard EM, and each iteration of EM($\eta$) requires exactly the same amount of computation as EM.

Our work leads to several interesting directions for future work. It is possible to modify the EM update rule to find the a (local) MAP (maximum a posteriori) parameter assignment. It should be fairly easy to similarly adapt the EM($\eta$) rule. It is also important to test the convergence of EM($\eta$) from a wider variety of random starting point, verifying that, in practice, it does converge to a local maximum from any starting point. If it does not, one could consider hybrid algorithms that adapted $\eta$ during the algorithm, using iterations with $\eta$ close to 1 to bring the algorithm close to a local maximum, and iterations with large $\eta$ to speed up convergence in the neighborhood.

A somewhat larger scale project is a more comprehensive investigation of the on-line learning task. We would like to prove convergence properties for the on-line update rules derived in Section 6, and to investigate their performance in practice. The evaluation metrics for this task are not so well-established, but we believe the on-line learning task to be a very important one in practice. Finally, we would like to investigate the applications of our basic framework, and of other recent results in on-line learning, to the task of learning Bayesian network structure.

### Acknowledgements

We would like to thank Nir Friedman, Yann Le Cun, Kevin Murphy, Manfred Warmuth, and the anonymous referees for useful comments and discussions. The work of Eric Bauer and Daphne Koller was supported by Silicon Graphics Inc. and by ONR grant N00014-96-1-0718. Some of the work was done while Daphne Koller was visiting AT&T Research. The experiments described here were done using $\mathcal{MLC}$++.[4]

---

[4] Available from http://www.sgi.com/Technology/mlc/.